\documentclass[conference]{IEEEtran}
\usepackage{times}

\usepackage[numbers]{natbib}
\usepackage{multicol}
\usepackage[bookmarks=true]{hyperref}

\usepackage{graphicx}
\usepackage{amsmath}
\usepackage{amssymb}
\usepackage{booktabs}
\usepackage{algorithm}
\usepackage{algpseudocode}
\usepackage{xspace}

\usepackage{amsmath,amsfonts,bm}

\def\eqref#1{equation~\ref{#1}}

\def\1{\bm{1}}

\def\va{{\bm{a}}}

\def\vc{{\bm{c}}}

\def\ve{{\bm{e}}}

\def\vj{{\bm{j}}}

\def\vp{{\bm{p}}}

\def\vu{{\bm{u}}}
\def\vv{{\bm{v}}}

\def\vx{{\bm{x}}}

\def\mA{{\bm{A}}}
\def\mB{{\bm{B}}}

\def\mD{{\bm{D}}}

\def\mI{{\bm{I}}}

\def\mK{{\bm{K}}}

\DeclareMathAlphabet{\mathsfit}{\encodingdefault}{\sfdefault}{m}{sl}
\SetMathAlphabet{\mathsfit}{bold}{\encodingdefault}{\sfdefault}{bx}{n}

\def\sN{{\mathbb{N}}}

\def\sR{{\mathbb{R}}}

\def\sZ{{\mathbb{Z}}}

\newcommand{\normlone}{L^1}
\newcommand{\normltwo}{L^2}
\newcommand{\normlp}{L^p}

\usepackage{amsthm}
\usepackage{tablefootnote}

\newtheorem{definition}{Definition}

\makeatletter
\DeclareRobustCommand\onedot{\futurelet\@let@token\@onedot}
\def\@onedot{\ifx\@let@token.\else.\null\fi\xspace}

\def\ie{\emph{i.e}\onedot}

\makeatother

\usepackage[capitalize]{cleveref}
\crefname{section}{Sec.}{Secs.}
\Crefname{section}{Section}{Sections}
\Crefname{table}{Table}{Tables}
\crefname{table}{Tab.}{Tabs.}
\crefname{algorithm}{Algo.}{Algos.}
\crefname{equation}{Eq.}{Eq.}

\begin{document}

\title{CCIL: Context-conditioned imitation learning for urban driving}





\author{\authorblockN{Ke Guo\authorrefmark{1} \authorrefmark{2}
Wei Jing\authorrefmark{2},
Junbo Chen\authorrefmark{2}, 
Jia Pan\authorrefmark{1}}
\authorblockA{\authorrefmark{1}The University of Hong Kong, Email: \{kguo, jpan\}@cs.hku.hk}
\authorblockA{\authorrefmark{2} Alibaba Group, Email: 21wjing@gmail.com, chjb@zju.edu.cn}}

\maketitle

\begin{abstract}
Imitation learning holds great promise for addressing the complex task of autonomous urban driving, as experienced human drivers can navigate highly challenging scenarios with ease. While behavior cloning is a widely used imitation learning approach in autonomous driving due to its exemption from risky online interactions, it suffers from the covariate shift issue. To address this limitation, we propose a context-conditioned imitation learning approach that employs a policy to map the context state into the ego vehicle's future trajectory, rather than relying on the traditional formulation of both ego and context states to predict the ego action. Additionally, to reduce the implicit ego information in the coordinate system, we design an ego-perturbed goal-oriented coordinate system. The origin of this coordinate system is the ego vehicle's position plus a zero mean Gaussian perturbation, and the x-axis direction points towards its goal position. Our experiments on the real-world large-scale Lyft and nuPlan datasets show that our method significantly outperforms state-of-the-art approaches. 
\end{abstract}

\IEEEpeerreviewmaketitle

\section{Introduction}

Planning a safe, comfortable, and efficient trajectory for a self-driving vehicle (SDV) in complex urban environments is a challenging and critical task in autonomous driving~\citep{yurtsever2020survey}. Unlike highway driving~\citep{henaff2019model}, urban driving requires handling various road geometries, such as roundabouts and intersections, while interacting with traffic lights, pedestrians, and other vehicles. Conventional rule-based approaches~\citep{fan2018baidu} have achieved some success in industry but require extensive human engineering to deal with diverse real-world scenarios. Recent advances in deep learning techniques have motivated researchers~\citep{bojarski2016end,pan2020imitation} to employ neural networks to model complex driving policies. Imitation learning (IL) from human drivers' demonstrations is a promising solution for learning these policies, as experienced drivers can handle even the most difficult situations, and their driving data can be easily collected at scale.

The simplest IL algorithm is the behavior cloning (BC) method, which has wide applications in autonomous driving~\citep{pomerleau1988alvinn,bojarski2016end,codevilla2018end}. It learns a policy in a supervised fashion by minimizing the difference between the actions taken by the learner and those taken by an expert in the expert state distribution without potentially dangerous online interactions. Despite its simplicity, the BC method suffers from the \textit{covariate shift} issue~\citep{ross2011reduction}, \ie the state induced by the learner's policy cumulatively deviates from the expert's distribution. 

To overcome the \textit{covariate shift} problem, existing methods such as DAgger~\citep{ross2011reduction} and DART~\citep{laskey2017dart} query supervisor corrections at the learner's or perturbed expert's states. Since human supervision is hard to collect, recent works like GAIL~\citep{ho2016generative} seek to provide feedback from a neural network-based discriminator to recover from out-of-distribution states induced by the learner's policy. However, these data augmentation methods need either expert supervision or rolling out the learner's policy in the real world, which is impractical in autonomous driving. Alternatively, some researchers have attempted to constrain the learned policy formulation to ensure its robustness to policy errors by incorporating control theoretic prior knowledge, as real-world systems usually have a robustness property. For example,~\citet{palan2020fitting,havens2021imitation} impose Kalman or linear matrix inequality constraints on learned linear policies to ensure closed-loop stability in a linear time-invariant system. \citet{yin2021imitation} extends this methodology by formulating the policy as a simple feed-forward neural network. Furthermore, \citet{east2022imitation} expands the method proposed in~\citep{havens2021imitation} to include polynomial policies and dynamical systems. However, the urban driving task is too complex to be handled by these naive linear or polynomial policy formulations. 

To learn a stable and general urban driving policy by imitating offline human demonstrations, we propose a \textbf{context-conditioned imitation learning} (CCIL) method. Unlike BC, our approach utilizes a policy network that predicts the SDV's future trajectory using only its context state, rather than taking both the ego and context state as inputs to generate the next action~\citep{codevilla2019exploring,bansal2018chauffeurnet}. In our method, the ego state represents the SDV's historical trajectory, while the context state encompasses the states of all other observed objects and the goal positions of the SDV. 

Our approach is primarily motivated by the fact that the ego state is highly susceptible to policy errors. Even a small distribution shift in policy inputs can result in greater action errors and eventually lead to out-of-distribution states. Besides, in the autonomous driving task, static elements in the context, such as lanes or crosswalks, remain unaffected by the SDV, while dynamic elements like human drivers attempt to recover from the SDV's perturbation. This stability property in the traffic system can be leveraged to address the distribution shift issue by considering only the context as policy input. In addition, removing the ego state from policy inputs can also help overcome the inertia problem caused by causal confusion~\citep{de2019causal}. For example, when the ego vehicle comes to a stop, the training data often shows a high probability of it remaining static. This leads to the formation of a spurious correlation between low speed and no acceleration, making it challenging to restart under the imitative policy. However, our method masks the ego history, thus preventing the formation of such spurious correlations.

However, simply removing the ego historical information from the policy inputs is not enough in practice. The policy inputs are usually transformed into the vehicle coordinate system, with the SDV's rear axle midpoint as the origin and orientation as the $x$-axis direction~\citep{bojarski2016end,scheel2022urban}. However, using such an ego-centric coordinate system can still implicitly leak ego information. This is because the observation distributions in the ego-centric coordinate can vary based on the choice of the coordinate origin, which planning policies can exploit to infer the ego information. An analogy with the non-inertial reference frame in physics can help in understanding this concept. In an ego-centric frame of reference, a moving subject (such as a car) that experiences acceleration can deduce some movement information about itself from local observations, thereby leaking ego motion information. To minimize such implicit leaked information, we develop an ego-perturbed goal-oriented coordinate system that is less influenced by ego motion. The origin of this coordinate system is the SDV's current position plus a zero mean Gaussian perturbation, and the $x$-axis direction points towards its goal position.

The main contributions of our paper can be summarized as follows:

1. We present a novel context-conditioned imitation learning method to address the \textit{covariate shift} issue in offline imitation learning. Our method learns a policy that predicts the SDV's future trajectory using only its context as input, with a robustness assurance based on the assumption of context stability.

2. We apply our approach specifically to urban driving by removing the explicit ego state information in policy input and proposing the ego-perturbed goal-oriented coordinate system to minimize implicit ego information in the coordinate system.

3. We verify the effectiveness of our method on the real-world large-scale urban driving datasets, Lyft~\citep{houston2020one} and nuPlan~\citep{Caesar2021nuplan}, achieving state-of-the-art performance benchmarks. The videos and code for our method can be found at \url{https://sites.google.com/view/contextconditionedil}.

\section{Related Work}

\subsection{Imitation learning for autonomous driving}

The objective of applying IL in autonomous driving is to teach an autonomous vehicle to drive by mimicking human drivers' behavior. The most straightforward approach is BC, which minimizes the difference between the learner's and expert's actions in the expert states without requiring additional manually labeled data or online interaction. Early BC applications in autonomous driving, such as ALVINN~\citep{pomerleau1988alvinn} and PilotNet~\citep{bojarski2016end}, used a large amount of human driving experience to learn an end-to-end policy that directly maps sensor inputs to vehicle control commands. Recently, ChauffeurNet~\citep{bansal2018chauffeurnet} improved generalization and transparency by yielding intermediate planning using perception results. However, the BC approach often leads to a \textit{covariate shift} between the training distribution and the deployment distribution. This means that even minor errors in the policy can cause the vehicle to deviate from the expert's state, leading to larger errors. To address this challenge, existing IL methods can be categorized into online methods, offline model-free methods, and offline model-based methods.

\subsubsection{Online IL}

Online methods aim to directly match the expert state-action distribution instead of matching the expert state-conditioned action distribution, as done in BC. For example, the method proposed in~\citep{Zhang2017query}, based on DAgger~\citep{ross2011reduction}, queries supervisor actions in the states visited by the learner and adds the new data into the dataset. This adjusts the expert state distribution to match the learner's state distribution, eliminating the requirement of an interactive expert. To get rid of the requirement of an interactive expert, methods like~\citep{wang2021decision} based on GAIL~\citep{ho2016generative} utilize a discriminator to measure the difference between the learner's and the expert's state-action distribution and then compute the reward in reinforcement learning. By increasing the policy's accumulated reward, the policy's state distribution will get closer to the expert distribution. However, deploying such reinforcement loops in safety-critical tasks like autonomous driving is challenging since it requires online interaction with the environment. In contrast, our proposed method matches the ego future trajectory's distribution conditioned on context, which can be learned through supervised learning without interaction with the environment or access to expert supervision. This avoids the need for online interaction with the environment and facilitates deployment in safety-critical scenarios such as autonomous driving.

\subsubsection{Offline Model-free IL}

The most popular model-free IL methods are based on DART~\citep{laskey2017dart} which avoids the compounding error by providing synthetic examples of how to recover from the deviated state. In~\citep{codevilla2018end}, temporally correlated noise is injected into the trajectory to simulate gradual drift away from the desired trajectory. Alternatively, ChauffeurNet~\citep{bansal2018chauffeurnet} adds a uniform perturbation to the SDV's current pose, \ie its rear axle’s midpoint coordinate and orientation, and fits a new smooth trajectory that brings the SDV back to the original target location. However, these rule-based trajectory augmentation methods are challenging to cover the real motion distribution induced by the learner's policy, and the policy is likely to develop a tendency for perturbed driving. In our approach, we also apply perturbation to the SDV's current position, but the perturbation's role is to blur the ego position information, rather than to provide data augmentation. Thus, our method does not require a trajectory smoothing process during training. Our method is also model-free, but we aim to endow the policy with robustness properties by constraining its formulation without relying on recovery examples.

\subsubsection{Offline Model-based IL}

To address the distribution shift problem, model-based IL methods minimize the difference between a trajectory rolling out in a differentiable learned or data-driven model and the expert trajectory. For example, PPUU~\citep{amos2018differentiable} learns a data-driven dynamics model based on a variational autoencoder~\citep{kingma2013auto} and trains the policy network to output actions that generate a similar trajectory as the expert trajectory. As the dynamics model is differentiable, an action can receive gradients from multiple time steps ahead, enabling the penalization of actions that will result in large divergences in the future, even if their instantaneous divergences are small. UrbanDriver~\citep{scheel2022urban}, instead of learning a model, constructs a differentiable data-driven model using recorded perception data and High Definition maps, where new observations are calculated by a coordinate transformation based on the ego vehicle's pose and collected data. However, the model-based approaches' performance is limited by their models' accuracy.

\subsection{Imitation learning with robustness}

IL is different from supervised learning by deploying the policy under dynamics, whose robustness is considered to be the learned policy's ability to recover from policy errors. Control-theoretic methods have been explored to learn policies with stability guarantees by constraining policy and system dynamics. Taylor Series IL~\citep{pfrommer2022tasil} shows that the induced trajectories of a learner and expert will be close if their derivative difference at expert states is small. However, computing high-order derivatives of the expert policy is difficult without sufficient data. Others learn a robust linear~\citep{palan2020fitting,havens2021imitation} or simple feed-forward neural network~\citep{yin2021imitation} control policy for a linear dynamical system by posing constraints on the policy. Even though the authors in~\citet{east2022imitation} extend the robustness guarantee to the polynomial system and policy, obtaining guarantees on close-loop stability for the nonlinear autonomous driving system remains a challenge. Recently, the CMILe method~\citep{tu2022sample} learns nonlinear policies with the same safety guarantees as the expert but requires online expert access, similar to DAgger~\citep{ross2011reduction}. In contrast, our method constrains our policy in the formulation of receiving context state and producing ego future trajectory, whose stability is guaranteed under a mild input-to-state stability assumption on the environment's dynamics.

\section{Theoretical Analysis}

We first introduce the notations and definitions used in this paper. For any vector $\vx \in \sR^n$, $\|\vx\|_p$ denotes its $\normlp$ norm, while $\| \vx \|$ represents $\normltwo$ norm. For any square matrix $\mA \in \sR^{n\times n}$, $\|\mA\|$ represents its induced $\normltwo$ norm and $\rho(\mA)$ represents its spectral radius (the maximum of the absolute values of its eigenvalues). For every induced matrix norm, we have $\rho(\mA)\leqslant \|\mA\|$.  

\begin{definition}[Comparison functions]
A function $\gamma: \sR_{\geqslant 0} \rightarrow  \sR_{\geqslant 0}$ is class $\mathcal{K}$ if it is continuous, strictly increasing and satisfies $\gamma(0)=0$. A function $\beta(x,t):\sR_{\geqslant 0} \times \sR_{\geqslant 0} \rightarrow  \sR_{\geqslant 0}$ is class $\mathcal{KL}$ if it is continuous, $\beta(\cdot,t)$ is class $\mathcal{K}$ for each $t$ and $\beta(x,\cdot)$ is decreasing for each $x$. 
\end{definition}

\begin{definition}[Input-to-state stability (ISS)~\citep{sontag2008input}] A discrete-time system $\vx_{t+1}=f(\vx_t,\vu_t), \vx_0=\bm{\xi}$ is input-to-state stable if there exists a class $\mathcal{KL}$ function $\beta$ and a class $\mathcal{K}$ function $\gamma$ such that, for each bounded input $\vu$ and initial condition $\bm{\xi}$ and $ t \in \sZ_+$, it holds that:
\begin{equation}
    \|\vx_t(\bm{\xi},\vu)\|\leqslant \beta(\|\bm{\xi}\|,t)+\gamma(\|\vu\|_\infty),
    \label{eq:iss}
\end{equation}
where $\|\vu\|_\infty=\sup_{t \in \sZ_+}(\vu_t)$ is the input's sup norm. If a system satisfies the ISS property, the distance between any two trajectories must eventually be bounded by its input signal and independent of initial conditions.
\end{definition}

We consider performing imitation learning in a nonlinear discrete-time system as follows:
\begin{equation}
    \vx_{t+1}=f(\vx_t,\vu_t),
\label{eq:nonlinear}
\end{equation}
where $\vx_t \in \sR^n$ is the system state at time $t\in \sN$ and $\vu_t \in \sR^u$ is the control input. For an autonomous driving system, its state can be separated into two parts $\vx_t=(\ve_t,\vc_t)$: the ego state $\ve_t \in \mathbb{R}^m$ controlled by the control input $\vu$ and the context state $\vc_t \in \mathbb{R}^{n-m}$ influenced by the ego state. By viewing the policy error as a disturbance to the system, we apply linearization to the nonlinear system at the expert's state $\vx^*$ to obtain the following linear subsystem:
\begin{equation}
    \vc_{t+1}=\mA\vc_t+\mB\ve_t, 
    \label{eq:linear}
\end{equation}
with solution: 
\begin{equation}
    \vc_{t+1}=\mA^{t+1} \vc_0+\sum_{j=0}^t \mA^{t-j} \mB \ve_j.
\label{eq:result}
\end{equation}
Here with abuse of notation, we use the same symbol $\vc_t$, $\ve_t$ for the context deviation $\vc_t - \vc_t^*$ and the ego state deviation $\ve_t - \ve_t^*$.

Because the real-world traffic system without the ego state deviation $\vc_{t+1}=\mA\vc_t$ is stable, we have $\rho(\mA)<1$. For such a Schur matrix, there are constants $c>0$ and $0 \leqslant \sigma<1$ such that $\|\mA^t\|\leqslant c\sigma^t$~\citep{jiang2001input}. Besides, the context in traffic systems generally includes static map elements that are not influenced by ego vehicles and intelligent human drivers who can quickly recover from the other vehicle's perturbation. It is natural to assume that the subsystem in~\cref{eq:linear} satisfying ISS property with $\beta(r,t) < \sigma^{t-1}r$ and $\gamma(r)\leqslant \epsilon r$, where $\epsilon\geqslant 0$. Combining~\cref{eq:iss} and~\cref{eq:result}, we have:
\begin{equation}
\begin{aligned}
\beta(r,t)&=c\sigma^t r < \sigma^{t-1} r, \\
\gamma(r)&=\sum_{j=0}^{\infty} c \sigma^t \|\mB\|r=\frac{c\|\mB\|}{1-\sigma}r \leqslant \epsilon r,
\end{aligned}   
\end{equation}
which implies that $c < \frac{1}{\sigma}$, and $\|\mB\|\leqslant \frac{\epsilon(1-\sigma)}{c}$.  

Next, we study how the policy in our formulation can be stable under this ISS system. We learn a context state feedback control $\ve_t=\vu_t=f_{\bm{\theta}}(\vc_t)$ that directly maps the context state to the ego state. For simplicity, we consider a linear policy $\vu_t=\mK \vc_t$. Then, the linear subsystem in~\cref{eq:linear} can be simplified as $\vc_{t+1}=(\mA+\mB\mK)\vc_t$ whose stable condition is $\rho(\mA+\mB\mK)<1$. So, we have when $\|\mK\|<\frac{c(1-c\sigma)}{\epsilon(1-\sigma)}$,
\begin{equation}
\begin{aligned}
\rho(\mA+\mB\mK) &\leqslant\|\mA+\mB\mK\|\leqslant\|\mA\|+\|\mB\|\|\mK\| \\
&\leqslant c\sigma+\frac{\epsilon(1-\sigma)}{c}\|\mK\|<1.
\end{aligned}
\label{eq:norm}
\end{equation}
This result shows that the stability of closed-loop autonomous driving can be guaranteed as long as the context system is stable enough and the norm of our policy is sufficiently small. In practice, we will add a $\normltwo$ norm regularization to the training loss of our neural network policy to internalize this stability prior.

Finally, we show how even with a fixed context $\vc_t=0$, the naive BC policy formulation, which takes both the ego state and the context state and outputs the displacement from the current state, can fail. We consider a linear policy: $\vu_t=\mK_{bc} \ve_{t}$ and update the state $\ve_{t+1}=\ve_t+\vu_t$. Then, the dynamics is $\ve_{t+1}=(\mI+\mK_{bc})\ve_t$. Because $\|\mI\|=1$, we cannot obtain the stable condition $\rho(\mI+\mK_{bc})<1$ by limiting the $\|\mK_{bc}\|$ as our policy formulation. Thus, the system's stability cannot be guaranteed.

This simple analysis supports our decision to disregard the SDV's historical trajectory during the learning process to avoid the \textit{covariate shift}. Its advantage will also be validated in our experiments, showing that our method can significantly decrease the off-road rate. 

\section{Method}

\begin{figure*}[t]
	\centering
	\includegraphics[width=\linewidth]{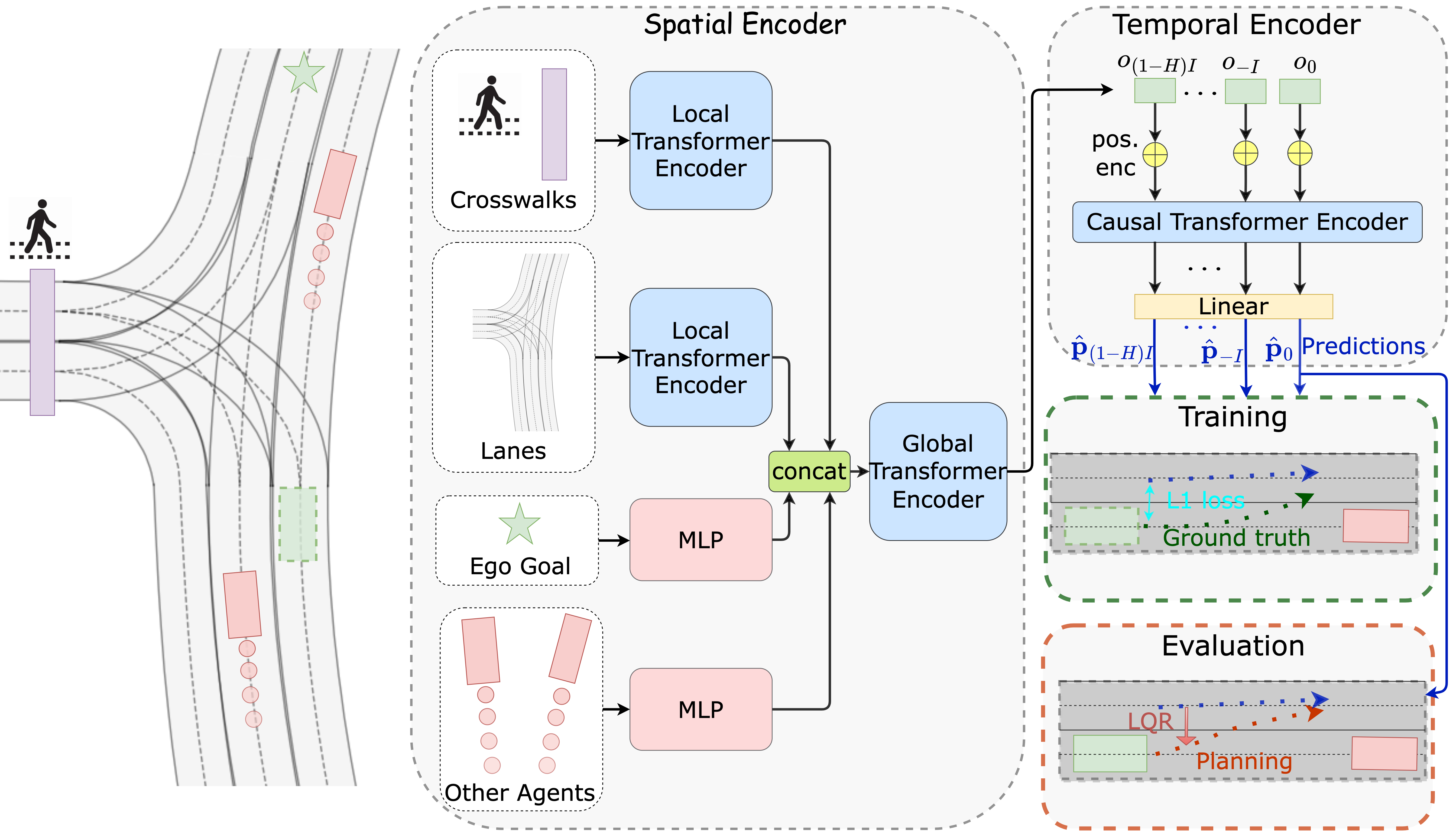}
	\vspace{-0.7cm}
	\caption{Overview of our approach.}
	\label{fig:overview}
\end{figure*}

In \cref{fig:overview}, we present an overview of our context-conditioned imitation learning method. Our policy network comprises two encoders, namely a spatial encoder and a temporal encoder based on the Transformer~\citep{vaswani2017attention}. During training, we optimize the network parameters to minimize the $\normlone$ distance between its predicted trajectories and the ground truth trajectories. During the evaluation, we employ a linear-quadratic regulator (LQR)~\citep{aastrom2021feedback} controller to yield smooth planning based on the predicted trajectory from the policy network. 

\subsection{Policy Inputs} 

Our policy network takes observed contexts from $H$ previous time steps with a time step interval of $I$ as inputs. Instead of directly learning an end-to-end policy from raw sensor data, such as camera images or lidar point clouds, we utilize intermediate input representations to improve generalization and interpretability, similar to the approach in~\citep{bansal2018chauffeurnet}.

The road networks consist of crosswalks and lanes, each represented as a polyline sequence of vectors with various features like VectorNet~\citep{gao2020vectornet}. Crosswalk vector features include their initial and terminal points' positions and sequence orders, while lane vectors also contain additional features such as their traffic light state and lane width.

Regarding the traffic participants, we divide them into the ego vehicle and other agents. As previously mentioned, we ignore the features of the ego vehicle that can be influenced by the learned policy and only consider its mission goal position without considering its historical trajectory. For other agents, we input their type, sizes, centroids, and orientations of bounding boxes at multiple past time steps.

Overall, our approach uses intermediate vector representations and disregards the SDV's historical trajectory to improve model performance.

\subsection{Coordinate System}

After obtaining contexts without the explicit ego state for $H$ time steps, we translate each context into its own current coordinate system, allowing for feature reuse during closed-loop evaluation. To avoid leaking implicit ego information that could be detrimental to closed-loop performance, we develop an ego-perturbed goal-oriented coordinate system. The origin of this coordinate system is anchored at the ego vehicle's position plus a zero-mean Gaussian perturbation, and the $x$-axis points to its goal position. The perturbation blurs the ego vehicle's position information and ensures local observability. The goal position eliminates the influence of the SDV's heading, which can be replaced with any point unaffected by the learned policy. 

It's worth noting that we apply the perturbation independently at all past $H$ time steps to blur or conceal the implicit ego position information from the learned policy, rather than to augment training data. This is because perturbed data is nearly impossible to occur during training and evaluation. In contrast, ChauffeurNet~\cite{bansal2018chauffeurnet} jitters the SDV's current pose, fits a smooth trajectory to the perturbed point and the original start and end points, and generates realistic new training examples.

\subsection{Policy Network} 

The policy network is designed to predict the SDV's future trajectory using only its context. Since predicting the SDV's future trajectory without its historical trajectory is more challenging, we need to make full use of spatial-temporal interactions between the SDV and its surroundings. To achieve this, we use a spatial encoder to embed observations at each time step into an observation feature. We then use a temporal encoder to embed the spatial observation features and generate predictions. The Transformer serves as the foundation for the network, as it has shown exceptional performance in autonomous driving~\cite{scheel2022urban,ngiam2022scene}.

\textbf{Spatial Encoder:} To capture spatial relations between vectors of the same map polyline, we use a local Transformer encoder as the vector-vector interaction encoder. Next, we aggregate the features of vectors belonging to the same polyline by max-pooling to obtain polyline-level features. Additionally, we utilize two multi-layer perceptions (MLPs) to obtain agent and goal features. Furthermore, to model high-order interactions between the goal, agents, and map elements, we employ a global transformer encoder, with the goal embedding being output as observation features $o_t$.

\textbf{Temporal Encoder:} To capture temporal information and interaction, a Transformer encoder with a causal self-attention mask is harnessed to embed the $H$ historical observation features. Finally, each hidden state of the Transformer is decoded by a full-connected linear layer to generate a prediction of the SDV's poses for its relative future $T$ time steps. 

\subsection{Training Process}

We leverage offline supervised learning like BC to train the policy network by minimizing the $\normlone$ distance between the predictions and ground truth trajectories. This is because the $\normlone$ metric is more correlated with driving performance compared to the commonly used mean square error~\citep{codevilla2018offline}. To help the network converge and generalize, we additionally introduce an auxiliary task that minimizes the $\normlone$ error at all previous time steps in the causal Transformer and apply a squared $\normltwo$ norm regularization to the network parameters $\bm{\theta}$, as inspired by~\cref{eq:norm}. The final loss is: 
\begin{equation}
    L=\sum_{t=1}^{T} \left( \|\vp_{t}-\hat{\vp}_{0}^{t}\|_1+\mu \sum_{h=1}^{H-1} \|\vp_{t-hI}-\hat{\vp}_{hI}^{t}\|_1\right)+\frac{\lambda}{2} \|\bm{\theta}\|^2,
\end{equation}
where $\vp_{t}$ denotes the ground truth ego pose at time step $t$
and $\hat{\vp}_{hI}^{t}$ is the pose prediction at the past $hI$ time step for its relative future $t$ time step. $\mu$ is an auxiliary task hyperparameter, and $\lambda$ is a regularization factor. 

\subsection{Evaluation Process }

While the prediction without ego state input is more stable, it struggles to ensure the smoothness of the predicted trajectory from the current state. To address this, prior works~\citep{vitelli2022safetynet,amos2018differentiable} have typically added a differentiable kinematic layer into the policy network to generate physically feasible planning. However, this differentiable kinematic layer would incorporate the SDV's current information, which is undesirable in learning a policy. Instead, we choose to obtain a smooth trajectory during the evaluation process by applying the LQR.

LQR is a computationally efficient method that minimizes the total commutative quadratic cost of a linear dynamic system. For simplicity, we consider a finite-horizon, discrete-time linear system with dynamics described by:
\begin{equation}
\left[\begin{array}{c}
    \vp_{t+1} \\
    \dot {\vp}_{t+1} \\
    \ddot{\vp}_{t+1}
    \end{array}\right] 
    =\left[\begin{array}{ccc}
            \mI & \mD & \mD^2 \\
            0 & \mI &  \mD \\
            0 & 0 & \mI
            \end{array}\right] 
            \left[\begin{array}{c}
                \vp_t \\
                \dot {\vp}_{t} \\
                \ddot{\vp}_{t}
                \end{array}\right] 
            + \left[\begin{array}{c}
                \mD^3 \\
                \mD^2 \\
                \mD
                \end{array}\right] \vu_t,
\end{equation}
where $\mD$ is a diagonal matrix with the interval of each time step as diagonal entries, $\dot{\vp}=(\omega_t,\vv_t),~\ddot{\vp}_{t}=(\alpha_t,\va_t),~\vu_{t}=(\zeta_t,\vj_t)$ represent the angular and positional velocity and acceleration, jerk, and control input, respectively. The system is subject to a quadratic cost function:
\begin{equation}
    J=\sum_{t=1}^T \|\vp_t-\hat{\vp}_0^t\|^2+\eta_\omega \omega_t^2+\eta_{\alpha}\alpha_t^2+\eta_{\va}\|\va_t\|^2+\eta_{\vj}||\vj_t||^2,
\end{equation}
where $\vp_t$ represents the planned pose, and the predicted pose $\hat{\vp}_{0}^{t}$ from the policy network are regarded as target pose. We use weights $\eta_\omega$, $\eta_{\alpha}$, $\eta_{\va}$and $\eta_{\vj}$ to balance safety and smoothness of the planned trajectory. At the end of the optimization, a smooth trajectory with positions and headings can be generated for the SDV to follow.

\section{Experiments}
\subsection{Dataset}

To evaluate the performance of our method, we conduct experiments on two large-scale real-world datasets:

\textbf{Lyft Level 5 Prediction Dataset}~\citep{houston2020one}: It contains approximately 1,000 h of urban driving demonstrations from Palo Alto, which have been separated into independent scenes of nearly 25 s at a frequency of 10 Hz. We train our network on the provided 100 h subset (16,265 scenes) as UrbanDriver~\citep{scheel2022urban} and test it with all 16,220 validation scenes.

\textbf{nuPlan Dataset}~\citep{Caesar2021nuplan}: This dataset consists of 1,312 h of human driving data collected in four cities (Boston, Pittsburgh, Las Vegas, and Singapore). Due to significant differences in traffic rules and patterns between different cities, we extract driving data in Las Vegas as~\citet{phan2022driving}. Then, we segment the data into independent scenes of 25 s at a frequency of 10 Hz, as in the Lyft dataset, and filter out scenes without a mission goal. After filtering, we obtain 63,181 training, 4,774 validation, and 6,386 testing scenes. 

More information on data preparation and another \textbf{toy Dataset} can be found in the appendix.

\subsection{Closed-Loop Evaluation}

To evaluate the closed-loop performance of our method, we use a log-replay simulator, as in prior work~\citep{scheel2022urban, huang2022differentiable}. At each step in the log-replay simulator, the SDV updates its pose according to the planned trajectory, while other agents are assumed to follow their recorded trajectories in the dataset. For both datasets, we evaluate our method for 25 s at a frequency of 10 Hz, using the following metrics:

\textbf{Collision Rate}: If the SDV collides with other agents at any time step during a scene, that scene is considered a collision scene. The collision rate is calculated by taking the ratio of the number of collision scenes to the total number of scenes.

\textbf{Off-road Rate}: In nuPlan, we use the official off-road metric, which considers the distance between a corner of the SDV's bounding box and the drivable area. Specifically, if this distance exceeds 0.3 m, the scene is deemed off-road. However, in the Lyft dataset, there is no access to the drivable area. Thus, following UrbanDriver~\citep{scheel2022urban}, we consider a scene off-road if the SDV deviates laterally from the human driver's ground truth by more than 2 m in the scene.

\textbf{Discomfort}: To quantify the comfort and feasibility of the planned trajectory, we calculate the rate at which the absolute acceleration values exceed 3 $\mathrm{m/s^2}$ across all time steps.

\textbf{L2}: We use the average L2 position errors between the roll-out trajectory and the human driver's ground truth to quantify the similarity between our results and human driving.

\subsection{Performance Evaluation}

On the Lyft dataset, we compare our methods against three state-of-the-art methods to demonstrate the advantage of the proposed framework:

\textbf{Raster-perturb}: An official baseline learned by BC for the Lyft dataset~\citep{houston2020one}, based on ResNet-50~\cite{resnet50}, which receives a Bird-Eye-View (BEV) representation of the scene surrounding the SDV and plans a trajectory with position and yaw displacements. To augment the data, a perturbation is applied to the current SDV position, and then a new kinematically feasible trajectory to reach the original endpoint is generated as ChauffeurNet.

\textbf{BC-perturb}: A BC method provided by UrbanDriver with the same trajectory perturbation and output as \textbf{Raster-perturb} but its map and agent inputs are represented in vector formulations, which are processed by a VectorNet and Transformer. In addition, the SDV's history is equipped with a dropout.

\textbf{UrbanDriver}: An offline policy gradient method to imitate the expert's policy by exploiting a differentiable data-driven simulator using the same input and model structure as \textbf{BC-perturb}.

On the nuPlan dataset, we compare our methods against an \textbf{official} baselines learned by BC due to a lack of other prior works:

\textbf{LaneGCN-perturb}: A vector-based model that utilizes a series of MLPs to encode SDV and agent inputs and a LaneGCN~\citep{liang2020learning} to encode vector-map elements, and then a fusion network to capture lane and agent intra- and inter-interactions through attention layers. To augment the training data, the SDV trajectories are perturbed and other agents are randomly dropped out. 

In addition to these models provided by prior works, we also learn our model using a representative method in behavior cloning and offline reinforcement learning on both datasets:

\textbf{TD3+BC}~\citep{fujimoto2021minimalist}: It adds a BC term to the policy updating of the Twin Delayed Deep Deterministic Policy Gradient (TD3)~\citep{td3} for implicit policy constraints. We construct the offline RL dataset by applying the trajectory perturbation augmentation and consider collision, and comfort for the reward design. 

\textbf{Vector-Chauffeur}: Our network learned using the same data augmentation method as the ChauffeurNet including trajectory perturbation and ego past motion dropout. And we represent the data in the same coordinate system, which uses the SDV's location as the origin and its heading perturbed by a uniform noise as the orientation. 

Our proposed method outperformed all previous work significantly on both datasets, as shown in~\cref{tab:result}. The nuPlan dataset had a higher collision rate than the Lyft dataset due to the scenarios being more complex with more road agents. The poor performance of the official baseline \textbf{LaneGCN-perturb} on the nuPlan dataset is due to using the official hyper-parameters without fine-tuning, as its training process is time-consuming.

\begin{table*}[t]
\caption{Comparison with baselines of closed-loop performance on the Lyft and nuPlan datasets. }
\begin{center}
\begin{tabular}{lccccc}
\multicolumn{1}{c}{\bf Model} & \multicolumn{1}{c}{\bf Num params}  &\multicolumn{1}{c}{\bf Collision(\%)} & \multicolumn{1}{c}{\bf Off-road(\%)} & \multicolumn{1}{c}{\bf Discomfort(\%)} & \multicolumn{1}{c}{\bf L2(m)} 
\\ \hline \\
Raster-perturb$^*$ & 23.6M & 15.48 & 5.06 & \textbf{4.00}  &  5.90\\
BC-perturb$^*$ & 1.8M & 9.38 &  6.77 & 39.10 &  4.77 \\
UrbanDriver$^*$ & 1.8M &  13.28 &  7.27 & 39.41 &  5.74 \\
TD3+BC  & 2.8M & 22.53$\pm$1.76  & 15.21$\pm$0.97   & 4.86$\pm$0.47 & 6.34$\pm$0.41   \\
Vector-Chauffeur & 1.5M &  10.12$\pm$0.23 & 3.40$\pm$0.32 & 5.42$\pm$0.44 & 5.03$\pm$0.43     \\
\textbf{CCIL (ours)} & \textbf{1.5M} &\textbf{3.32}$\pm$0.15 & \textbf{0.62}$\pm$0.13 & 4.33$\pm$0.22  & \textbf{1.23}  $\pm$ 0.08    \\
\midrule
LaneGCN-perturb & 2.0M & 60.63$\pm$2.34  & 34.25$\pm$1.65 & 17.26$\pm$1.80 & 21.21$\pm$1.81   \\TD3+BC & 2.8M & 39.12$\pm$2.21  & 18.59$\pm$1.04 & 10.56$\pm$0.95 &  15.04$\pm$1.62  \\
Vector-Chauffeur & 1.5M & 24.12$\pm$1.37 & 10.11$\pm$0.62 & 12.53$\pm$1.17 & 6.12$\pm$0.87  \\
\textbf{CCIL (ours)} & \textbf{1.5M} &\textbf{6.91}$\pm$0.11 & \textbf{3.08}$\pm$0.11 & \textbf{1.16}$\pm$0.05 & \textbf{3.68}$\pm$0.04 \\
\end{tabular}
\end{center}
\label{tab:result}
\footnotesize{$^*$There is no variance in Raster-perturb, BC-perturb, and UrbanDriver because we evaluate the deterministic pre-trained models in a deterministic simulator.}
\end{table*}

\subsection{Ablation Study}

The following ablation experiments on the Lyft dataset are used to expose the significance of different components of our model, whose results are shown~\cref{tab:ablation}:

\begin{table*}[t]
\caption{Ablation experiments on closed-loop performance on the Lyft dataset}
\begin{center}
\begin{tabular}{lcc|cccc}
\multicolumn{1}{c}{\bf Model} & \multicolumn{1}{c}{\bf Perturb} & \multicolumn{1}{c}{\bf Ego}  &\multicolumn{1}{c}{\bf Collision(\%)} & \multicolumn{1}{c}{\bf Off-road(\%)} & \multicolumn{1}{c}{\bf Discomfort(\%)} & \multicolumn{1}{c}{\bf L2(m)}
\\ 
\hline  \\
w explicit ego & \checkmark & \checkmark & 20.29$\pm$0.88 & 19.18$\pm$2.98 & \textbf{0.57}$\pm0.15$ & 5.50$\pm$0.55 \\
w ego dropout & \checkmark & \checkmark & 14.05$\pm$1.53 & 5.02$\pm$0.88 & 0.63$\pm$0.20 & 4.15$\pm$0.47  \\
\midrule
w ego coordinate &  & \checkmark & 11.31$\pm$1.44 & 9.79$\pm$1.34 & 0.95$\pm$0.05 & 3.87$\pm$0.11 \\
std=0  &  & & 7.08$\pm$0.35 &  2.86$\pm$0.25 & 0.89$\pm$0.10 & 3.46$\pm$0.34 \\
std=1 & \checkmark &  & 3.39$\pm$0.17 & 1.00$\pm$0.16 & 1.99$\pm$0.15 & 2.10$\pm$0.05 \\
std=2 (ours) &  \checkmark & & \textbf{3.32}$\pm$0.15 & 0.62$\pm$0.13 & 4.33$\pm$0.22   & 1.23$\pm$0.08   \\
std=3 & \checkmark &  & 3.42$\pm$0.12 & \textbf{0.49}$\pm$0.10 & 7.35$\pm$0.26 & \textbf{0.91}$\pm$0.04 \\
\midrule
w/o causal Trans & \checkmark &   & 4.28$\pm$0.25  & 1.43$\pm$0.25 & 6.53$\pm$0.26 & 1.63$\pm$0.10 \\
w/o LQR  & \checkmark &   & 3.81$\pm$0.14   & 2.07$\pm$0.12 & 89.05$\pm$0.35 & 1.02$\pm$0.02 \\
\midrule
w/o regularization & \checkmark & & 4.07$\pm$0.16 & 1.05$\pm$0.14 & 4.96$\pm$0.30 & 1.23$\pm$0.09  \\
w/o auxiliary   &\checkmark & & 4.56$\pm$0.29 & 1.02$\pm$0.06 & 3.23$\pm$0.31 &  1.92$\pm$0.07  \\
\end{tabular}
\end{center}
\label{tab:ablation}
\end{table*}

\textbf{Network Input}: We first study the importance of removing explicit ego information from the network input. We consider two approaches to reintroducing explicit ego inputs. One is to directly input the SDV's past positions and then process them in the same way that the goal position is processed. The other one is to additionally introduce a dropout of 50\% at the ego input during training as ChauffeurNet. Note that our method is different from dropout methods because the dropout is only applied during training while we keep removing the ego state. We can observe that in both explicit ego information input ways, there is a steep drop in the collision and off-road rate and L2 distance due to the \textit{covariate shift} issue. 

\textbf{Coordinate System}: To analyze the impact of the implicit ego information in the coordinate system, we first consider replacing our ego-perturbed goal-oriented coordinate system with the ego-centric coordinate system in ChauffeurNet using orientation uniformly around the heading. We observe that the ChauffeurNet coordinate system leads to inferior closed-loop performance due to the implicit ego information. Furthermore, in our coordinate system, we add a Gaussian noise with a zero mean to the SDV's current position to obtain the origin. By increasing the standard deviation (std) of the Gaussian noise, we can reduce the implicit ego information. We observe that as the std increased, discomfort also increased, while the off-road rate and closed-loop L2 decreased. This suggests that although implicit ego information can improve instantaneous planning accuracy, it may worsen long-term closed-loop driving performance. We guess this is because during closed-loop driving, an auto-regressive model can generate temporal correlations in planning errors, leading to consistent bias in one direction that results in off-road events. However, temporally uncorrelated noise can cause slight oscillations around the expert trajectory but still enable successful driving. The phenomenon that offline instantaneous planning accuracy and actual long-term driving quality are weakly correlated was also noted by~\citet{codevilla2018offline}.

\textbf{Model Architecture}: To demonstrate the importance of the temporal information and the effectiveness of the causal Transformer in capturing the temporal interactions, we replace it with a MLP. Our findings suggest that the causal Transformer achieves better performance by integrating both spatial and temporal information. Moreover, we conducted an ablation study on the LQR module to examine its impact on comfort. We observed a significant decrease in comfort when the LQR mechanism was removed, as well as a small increase in collision and off-road rates. However, there was a slight improvement in the L2 metric. This highlights the importance of the LQR module in enhancing trajectory smoothness while acknowledging that the ego information introduced during evaluation can still impair closed-loop performance on the L2 metric.

\textbf{Loss Term}: To demonstrate the effectiveness of the additional loss terms, we remove the auxiliary and regularization loss term respectively. We find that both are beneficial to improving performance, although the improvement brought by norm regularization is limited, likely because our policy network has a modest parameter number and is well-regularized even without the loss term.

\subsection{Qualitative Analysis}

\begin{figure}[t]
\centering
\includegraphics[width=0.75\linewidth]{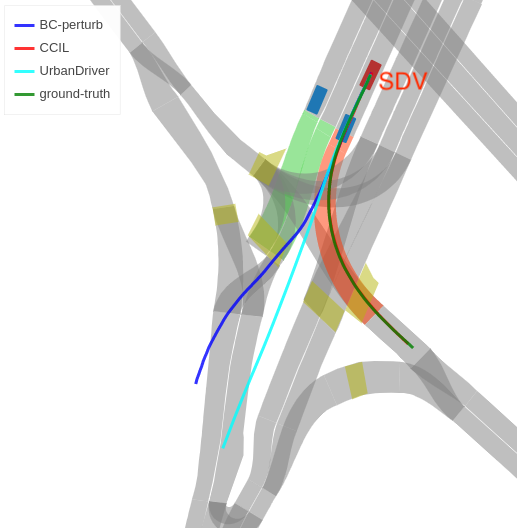}
\caption{Qualitative comparison between our method and the baselines for on-road driving.}
\label{fig:offroad}
\end{figure}

\begin{figure}[t]
\centering	\includegraphics[width=0.75\linewidth]{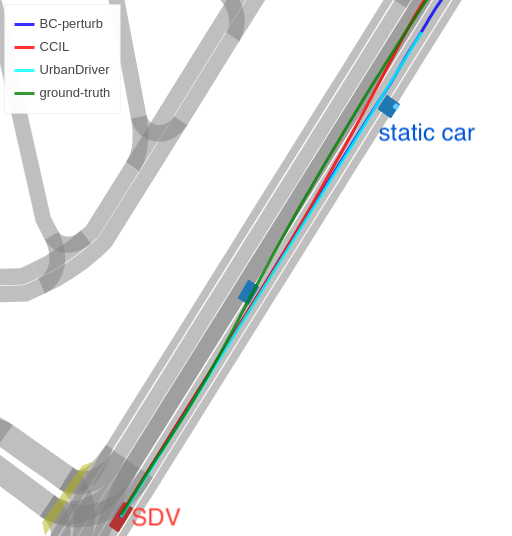}
\caption{Qualitative comparison between our method and the baselines for collision avoidance.}
\label{fig:collision}
\end{figure}

We visualize and compare the closed-loop trajectories of our method and the baselines in two Lyft scenarios. As shown in~\cref{fig:offroad}, our method generates a feasible trajectory similar to that of a human driver, whereas the BC-perturb and UrbanDriver methods drive the SDV offroad. We can see in~\cref{fig:collision} that our method can avoid the collision with the static car at the shoulder by changing lanes while the baseline methods keep moving straight, leading to a crash. We hypothesize that off-road or collision events occur when the policy outputs an erroneous action at the turning point or lane-changing point, such as continuing to move straight forward. In such cases, the policy taken in out-of-distribution history may make a larger error, leading to going off-road or colliding with other objects. The trajectory perturbation in BC-perturb and the trajectory distribution matching in UrbanDriver do not help the SDV recover from the erroneous state, which may be caused by the strong correlation between past straight trajectories and moving straight action in the training data.

\begin{figure}[t]
\centering
\includegraphics[width=0.75\linewidth]{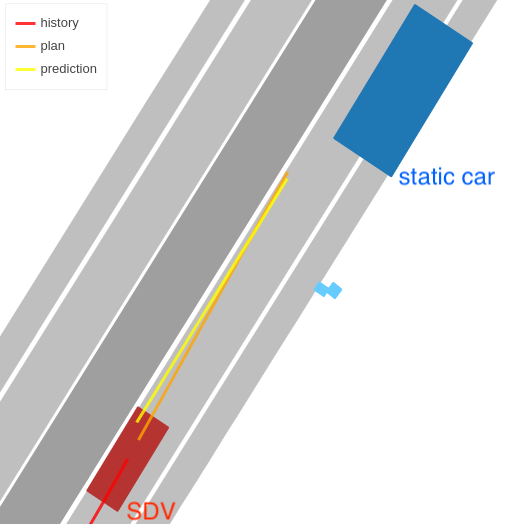}
\caption{Qualitative analysis of our model's prediction and planning performance. Please refer to~\cref{fig:collision} for the whole scenario.}
\label{fig:LQR}
\end{figure}

We also conducted a qualitative analysis of our model prediction and planning in the collision avoidance scenario of~\cref{fig:collision}. In~\cref{fig:LQR}, we demonstrate that our model can plan a smooth trajectory based on the policy network prediction. When faced with a static car ahead on the shoulder, our policy network infers that the SDV's future positions should be closer to the road center. Although the prediction is not dynamically feasible from the SDV's current state, it is close enough to it. This suggests that our policy can infer a reasonable ego state from the context.

\section{Conclusion}

We have proposed a new offline imitation learning method to mitigate the distribution shift of behavior cloning. Our approach involves training a neural network to predict the future poses of the ego vehicle, without receiving explicit ego information. To remove implicit ego information and ensure local observability, we introduce a new ego-perturbed goal-oriented coordinate system for representation of the observation input. To tackle the challenging learning task, we design a Transformer-based network that leverages historical spatial-temporal context data effectively. Finally, we demonstrate the effectiveness of our approach using two real-world large-scale datasets, achieving state-of-the-art performance.

\section*{Acknowledgments}
This work was supported by Alibaba Group through Alibaba Innovative Research Program and Alibaba Research Intern Program.

\bibliographystyle{unsrtnat}
\bibliography{mybib}

\newpage
\appendix

\subsection{Map prepossessing}

In Lyft, there are only two types of map elements: lane and crosswalk. In nuPlan, the map elements consist of a lane, lane connector, intersection, stop line, crosswalk, walkway, and car park. We categorize them into polyline elements including lanes and lane connectors, and polygon elements including stop lines, crosswalks, intersections, walkways, and car parks. 

For each polyline element, it is approximated by a sequence of vectors with an interval of 3 m. For both datasets, we prepossess the polyline map into a graph to take advantage of the topological information. We connect each vector with its nearest left, right, and next vector if any exist. The nearest left or right vector is the nearest vector of its reachable left or right polyline. And the distance between vectors is represented by the Euclidean distance between their starting points. After building the vector graph, we can compute the travel distance between any two vectors using Dijkstra's algorithm. For both datasets, the polyline vector has shared features, including coordinates of the starting and ending points, distances to the left and right vectors, distances to the mission goal, traffic light states (red or green), and the sequence order. However, lane vectors in Lyft have a lane width feature because the vectors approximate the computed middle lines of lanes, while the lane or lane connector vectors in nuPlan have a lane left and width feature because the vectors approximate an annotated baseline. In addition, the polyline vectors in nuPlan have an additional speed limit and type feature. 

For each polygon element, we also approximate it with a sequence of vectors. In the Lyft dataset, the crosswalk vectors are directly constructed by connecting the original sequential annotation coordinates. But we approximate each polygon with a fixed set of 20 vectors because the annotation point number is too large for elements like intersections. For both datasets, each polygon vector has features including coordinates of starting and ending points and its order sequence, while vectors in nuPlan have additional type features.

The inputs to our neural network are composed of two types of map elements: polylines and polygons, and two types of agent elements: other agents and an ego goal. The missing inputs are padded with zeros and masked out when calculating the attention. The origin is the perturbed SDV position. 

\textbf{Polyline}: 30 topologically nearest polylines with vectors whose starting points are within 35 m from the origin. The topological distance between the origin and a polyline is the minimum of the distances between the origin and its vectors.

\textbf{Polygons}: 20 polygons whose boundaries are within 35 m from the origin. If there are more than 20 polygons, they are selected according to the importance of their types: stop line, crosswalk, intersection, walkway, and car park. 

\textbf{Agents}: the 30 nearest agents whose oriented boxes' centroids are within 50 m from the origin. The agent features in Lyft include its centroid coordinates, yaws, shapes, types, and relative times in the past 2 and current steps. The nuPlan agents additionally have velocity features, as they are provided.

\textbf{Goal}: the $x, y$ coordinates of the SDV's mission goal of a scene. In Lyft, the mission goal is not provided, so we regard the ending point of the lane where the SDV locates at the last time step of the scene as the mission goal. In nuPlan, we directly take the provided mission goal at the last time step of the scene as the scene mission goal. 

\subsection{Model}

For both datasets. the same model architecture is used, whose hyper-parameters are listed in~\cref{tab:Hyper-parameters}. 

\begin{table}[t]
\caption{Hyper-parameters for both datasets}
\label{tab:Hyper-parameters}
\begin{center}
\begin{tabular}{lc}
\multicolumn{1}{c}{\bf  Hyper-parameter}  &\multicolumn{1}{c}{\bf Value}
\\ \hline \\
    Future steps $T$ &  15 \\
    All Transformers dropout &  0.1 \\
    All Transformers head number & 8 \\
    All Transformers hidden size &  128 \\

    Local Transformer layer number & 3\\
    Global Transformer layer number  &  6 \\
    Causal Transformer layer number & 3\\
    
    Causal Transformer history length $H$ &  15 \\
    Causal Transformer interval $I$&  2 \\
     Auxiliary weight $\mu$ & 0.3 \\
     Regularization weight $\lambda$ & 0.0001 \\
     LQR angular velocity weight $\eta_\omega$ & 0.1\\
     LQR angular acceleration weight $\eta_\alpha$ & 0.1\\
     LQR positional acceleration weight $\eta_\va$ & 0.1\\
     LQR positional jerk weight $\eta_{\vj}$ & 0.01\\
\\
\end{tabular}
\end{center}
\end{table}

\subsection{Training}

Our model is trained using the Adam optimizer with a learning rate of 0.0005, 10000 steps linear warm-up, $\beta=(0.9, 0.999)$, and batch size 128. We stop training after 30 epochs and select the model with the smallest validation collision rate for evaluation. We train all models except the pre-trained model in the Lyft dataset independently 3 times and then report the mean and std of their performances. 

\subsection{Evaluation}

For Lyft baselines, we directly evaluate the pre-trained model provided by the Lyft dataset and UrbanDriver. The Raster-perturb is the \textbf{model trained on train.zarr for 2 epochs} at \url{https://github.com/woven-planet/l5kit/blob/master/examples/planning/train.ipynb}. The BC-perturb and UrbanDriver are the \textbf{Open Loop with history} and the \textbf{Urban Driver} at \url{https://github.com/woven-planet/l5kit/blob/master/examples/urban_driver/train.ipynb}. 

We evaluate all Lyft models from the second time step to compute current velocity using position information as input to the LQR with zero assumed initial acceleration. 

For nuPlan baselines, we train the official models by ourselves using provided hyper-parameters and evaluate from the first time step because the velocity and acceleration information is provided.

\subsection{Runtime}

We conduct runtime experiments using a single Nvidia GeForce GTX 1080 GPU and an Intel i7-8700@3.2GHz CPU on the Lyft dataset. We measure the runtime of each method its mean and std over all time steps in an evaluation. The runtime results shown in~\cref{tab:Runtime} consider all components in each model including data-prepossessing, model inference, and control. We observe that our architecture can achieve higher data processing efficiency and medium model inference efficiency compared with other methods. Our method takes a longer total execution time due to the extra LQR control module which does not exist in prior works because they only focus on optimizing positional accuracy but not comfort. However, our approach can still be executed in real-time on this hardware. 

\begin{table*}[htbp]
\caption{Averaged runtime per frame of individual components for each method}
\label{tab:Runtime}
\begin{center}
\begin{tabular}{lcccc}
\multicolumn{1}{c}{\bf  Model}  & \multicolumn{1}{c}{\bf Data process (ms)} & \multicolumn{1}{c}{\bf Model inference (ms)} & \multicolumn{1}{c}{\bf Control (ms)} &  \multicolumn{1}{c}{\bf Total (ms)}
\\ \hline \\
Raster-perturb & 6.03$\pm$0.61  & \textbf{4.62}$\pm$0.16  & -& \textbf{10.65}$\pm$0.64 \\
BC-perturb & 6.69$\pm$0.72 & 6.78$\pm$5.26 &- & 13.47$\pm$5.65 \\
UrbanDriver &  6.33$\pm$1.41 & 12.78$\pm$8.80 &-& 19.11$\pm$9.23 \\
CCIL &  \textbf{4.92}$\pm$0.63 & 6.74$\pm$0.34 & 11.09$\pm$0.27 & 22.75$\pm$1.07 \\
\end{tabular}
\end{center}
\end{table*}

\subsection{Toy Experiment}

\begin{figure}[htbp]
\begin{center}
\includegraphics[width=0.8\linewidth]{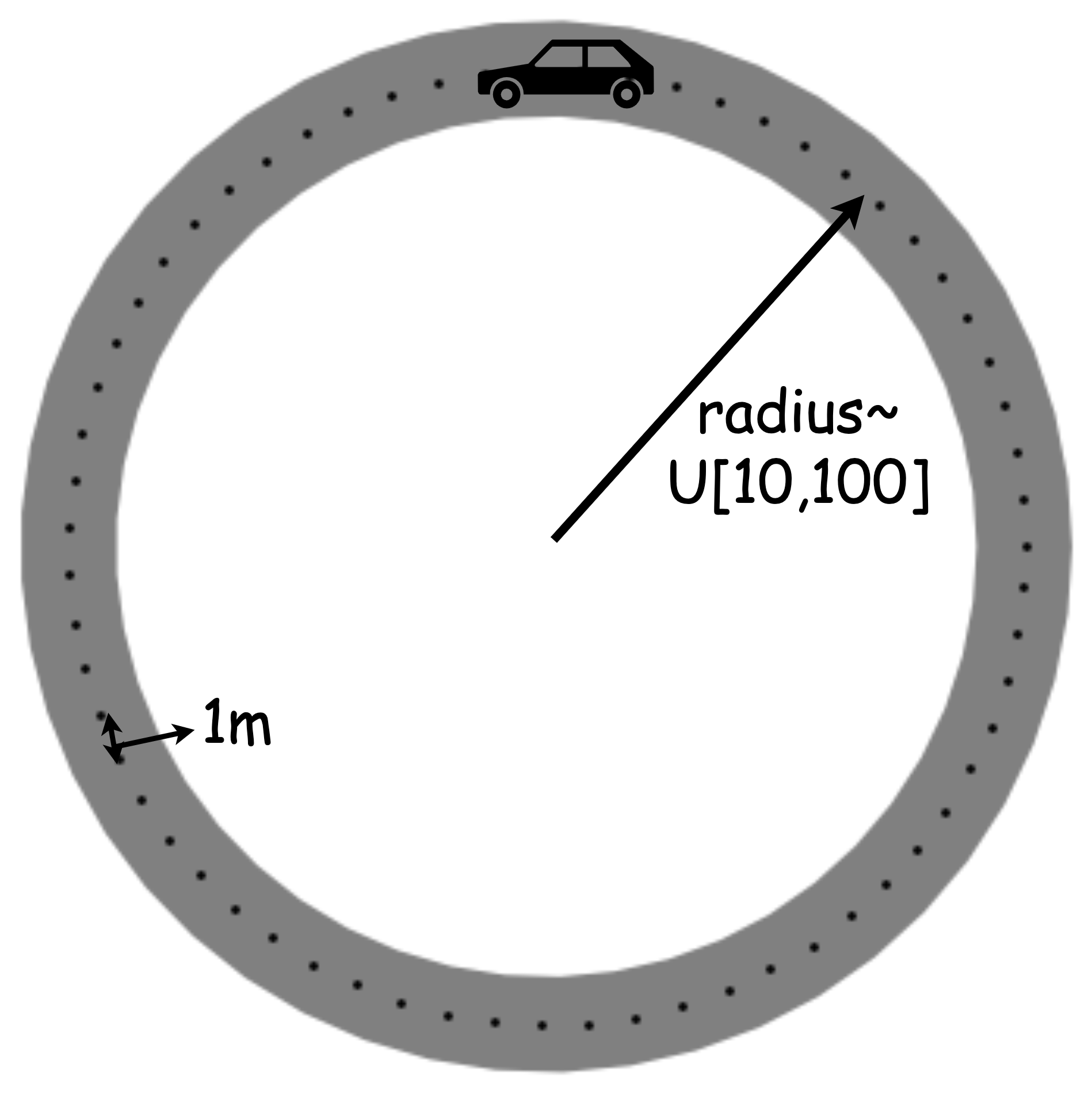}
\end{center}
\caption{Ring road}
\label{fig:ring}
\end{figure}

We design a toy experiment to vividly show our method's ability to reduce compounding error. In the experiment, we use synthetic data from a very ideal and simplified scenario where a SDV moves under a ring road network with a fixed speed of 1 m/s at 1 Hz, as shown in~\cref{fig:ring}. During training, the radius of the ring is a variable with a range from 10 m to 100 m and the circular lane is represented as a series of fixed lane points with the same interval of nearly 1m.

We compare our CCIL methods with several baseline methods introduced above, including BC, BC-perturb, and UrbanDriver. The inputs of these baseline methods are composed of two parts: ego state (its position and orientation in the past 10 time steps) and context (the nearest 10 lane points) in the ego coordinate system, while our method only takes the context in the past 10 time steps as inputs in the ego-perturbed center-oriented coordinate system. The ego-perturbed origin-oriented coordinate system means using the SDV position plus a zero-mean one-std Gaussian perturbation as the origin and orientation to the center of the circular road as the $x$-axis direction. The trajectory perturbation is applied to augment the data in the BC-perturb method. For outputs, the BC and CCIL methods generate the relative position and yaw at the next time step and the BC-perturb method produces the next 10 time steps. In the UrbanDriver method, we unroll the policy for 32 time steps. 

\begin{figure}[htbp]
\begin{center}
\includegraphics[width=\linewidth]{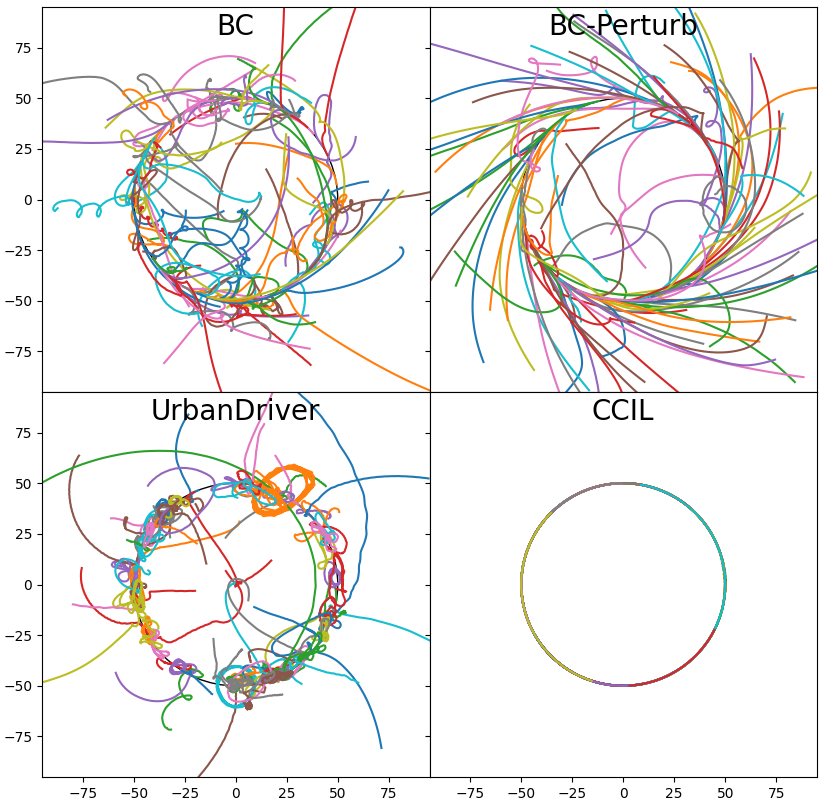}
\end{center}
\caption{Closed-loop trajectories of each model trained on the toy dataset.}
\label{fig:toy}
\end{figure}

In each method, we employ a two-layer MLP with a hidden size of 128 as a policy network. We train the neural networks using the Adam optimizer with a learning rate of 0.0001 and random initial weights 100 times. We stop training after 10000 steps and then unroll the policy from one random starting point on the circle of radius 50 m for 100 time steps. The 100 closed-loop trajectories for each method are depicted in~\cref{fig:toy}. We can observe that some trajectories in the baseline methods deviate from the road due to the covariate shift issue, but the trajectories in our method keep following the route. 

\end{document}